%% file: main_icra.tex
\documentclass[letterpaper, 10 pt, conference]{ieeeconf}  

\usepackage[bookmarks=true]{hyperref}
\usepackage{xcolor}
\hypersetup{
    colorlinks,
    linkcolor={red!80!black},
    citecolor={green!80!black},
    urlcolor={blue!80!black}
}

\IEEEoverridecommandlockouts                              
                                                          
\input{preamble.tex}

\DeclareMathOperator{\E}{\mathbb{E}}
\DeclareMathOperator{\T}{\mathcal{T}}

\newcolumntype{Y}{>{\centering\arraybackslash}X}
\DeclareCaptionLabelFormat{withbrackets}{#2}
\newacronym{rl}{RL}{Reinforcement Learning}
\newacronym{drl}{Dist. RL}{Distributional Reinforcement Learning}
\newacronym{gae}{GAE}{Generalized Advantage Estimation}
\newacronym{ppo}{PPO}{Proximal Policy Optimization}
\newacronym{sac}{SAC}{Soft Actor Critic}
\newacronym{ddpg}{DDPG}{Deep Deterministic Policy Gradients}
\newacronym{td3}{TD3}{Twin-Delayed DDPG}
\newacronym{d4pg}{D4PG}{Distributed Distributional DDPG}
\newacronym{dsac}{DSAC}{Distributional Soft Actor Critic}
\newacronym{lstm}{LSTM}{Long Short-Term Memory}
\newacronym{mlp}{MLP}{Multi-Layer Perceptron}
\newacronym{rnn}{RNN}{Recurrent Neural Network}
\newacronym{elu}{ELU}{Exponential Linear Unit}

\begin{document}
\title{Learning Risk-Aware Quadrupedal Locomotion using Distributional Reinforcement Learning}


\author{
  Lukas Schneider\\
  ETH Zurich \\
  \texttt{lukas@luschneider.com} \\
  \And
  Jonas Frey \\
  ETH Zurich \\
  \texttt{jonfrey@ethz.ch} \\
  \AND
  Takahiro Miki \\
  ETH Zurich \\
  \texttt{tamiki@ethz.ch} \\
  \And
  Marco Hutter \\
  ETH Zurich \\
  \texttt{mahutter@ethz.ch} \\
}

\author{Lukas Schneider$^{1}$, Jonas Frey$^{1,2}$, Takahiro Miki$^{1}$, and Marco Hutter$^{1}$
\thanks{$^{1}$ Robotic Systems Lab, ETH Zurich, Zurich, Switzerland}%
\thanks{$^{2}$ Max Planck Institute for Intelligent Systems, Tübingen, Germany.}%
\thanks{This work was supported by the Swiss National Science Foundation (SNSF) through project 188596, the National Centre of Competence in Research Robotics (NCCR Robotics), the European Union's Horizon 2020 research and innovation program under grant agreement No 101016970, No 101070405, and No 852044, and an ETH Zurich Research Grant. Jonas Frey is supported by the Max Planck ETH Center for Learning Systems.}
}

\maketitle

\begin{abstract}
Deployment in hazardous environments requires robots to understand the risks associated with their actions and movements to prevent accidents. Despite its importance, these risks are not explicitly modeled by currently deployed locomotion controllers for legged robots. In this work, we propose a risk sensitive locomotion training method employing distributional reinforcement learning to consider safety explicitly. Instead of relying on a value expectation, we estimate the complete value distribution to account for uncertainty in the robot's interaction with the environment. The value distribution is consumed by a risk metric to extract risk sensitive value estimates. These are integrated into Proximal Policy Optimization (PPO) to derive our method, Distributional Proximal Policy Optimization (DPPO). The risk preference, ranging from risk-averse to risk-seeking, can be controlled by a single parameter, which enables to adjust the robot's behavior dynamically. Importantly, our approach removes the need for additional reward function tuning to achieve risk sensitivity. We show emergent risk sensitive locomotion behavior in simulation and on the quadrupedal robot ANYmal. Videos of the experiments and code are available at \href{https://sites.google.com/leggedrobotics.com/risk-aware-locomotion}{https://sites.google.com/leggedrobotics.com/risk-aware-locomotion}.
\end{abstract}


\section{Introduction}

\begin{figure}[htb]
    \centering
    \includegraphics[width=\linewidth]{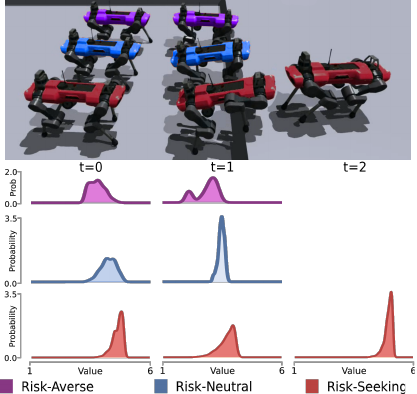}
    \caption{Our robot learns to adapt its locomotion behavior in risky situations. When commanded to walk up a large step, the risk-averse policy~(\magentasquare{}) refuses while the risk-seeking policy~(\redsquare{}) complies. The risk parameter, controlling the value distribution distortion, can be adapted online during deployment. Bottom: Respective risk-metric distorted value distribution per robot.}
    \label{fig:risk-adaptation-preview}
    \vspace{-0.8cm}
\end{figure}

Legged robots can traverse rugged terrain inaccessible to wheeled or tracked systems. They can overcome sloped and uneven terrain, stones, stairs, and even large gaps by carefully choosing their foot placement. This versatility makes them ideal for operation in hazardous environments, such as cave systems~\cite{nebula-spot,cerberus}, forests~\cite{forest1,forest2}, or the surfaces of other planets~\cite{legged-robots-gnc-space,legged-robots-scalability-space}. In these situations, the safe operation of the robot is paramount, as failure could result in hardware damage or mission failure.

Recent developments in model-free deep \acrfull{rl}~\cite{trpo, ppo} have enabled legged robots to traverse difficult terrain reliably~\cite{learning-quadrupedal-locomotion-over-challenging-terrain, learning-robust-perceptive-locomotion}. While these systems are robust against perception failures, a notable limitation remains: they do not account for risk explicitly.
Encoding behavior, such as avoiding dangerous obstacles or reducing locomotion speed on rough ground, requires adaptation of the reward formulation. Such adaptations are undesirable as learning a reliable locomotion already requires expensive and time-consuming reward function tuning.

In this work, we overcome this limitation by adapting the \gls{rl} algorithm to allow risk sensitive behavior to arise without changing the reward formulation.
We adopt the distributional perspective on \gls{rl}~\cite{c51}, learning the entire distribution of returns instead of just its scalar expectation. This value distribution models the intrinsic uncertainty of the agent's interaction with its environment. While estimating the full distribution was initially proposed for more accurate value estimates, it also captures useful information, such as the probability of catastrophic and high-return events. We use the value distribution in the actor-critic framework~\cite{d4pg} by incorporating risk sensitive value estimates into \acrfull{gae}~\cite{gae}. To extract these scalar value estimates, we employ a risk metric~\cite{robot-risk-metrics} that allows us to emphasize a specific part of the value distribution, e.g., unlikely but catastrophic events.
The learned locomotion policy accounts for risks explicitly. Since the metric encodes a fixed risk preference based on a metric parameter, and no single risk preference is appropriate under all circumstances, we condition the policy on this parameter~\cite{drl-worst-cases-pg}. Conditioning the policy allows the adjustment of the robot's behavior to either seek or avoid risks on demand by an operator or a high-level planner. For example, it enables the safe teleoperation of the robot by an amateur in risk-averse mode while recovering the full range of capabilities in risk-neutral / seeking mode. Figure~\ref{fig:risk-adaptation-preview} visualizes the behavior of different risk sensitivities of a single policy.

We thoroughly analyze the risk sensitive behavior arising from our method and study the effects of using different risk metrics and parameters in simulation. We further ablate the algorithm and deploy the trained policy to the quadrupedal ANYmal robot~\cite{anymal}, demonstrating risk sensitive behavior based on the risk parameter when climbing steps of varying heights.

Our contribution is threefold. Firstly, we integrate explicit risk modeling into the locomotion controller, a capability that currently deployed controllers do not possess. Secondly, we study the emergent risk sensitive locomotion behavior in simulation and on hardware. Lastly, on the algorithmic side, we explore introducing risk sensitivity into the \gls{rl} formulation through risk sensitive advantage estimates extracted from the value distribution.


\section{Related Work}
\label{sec:related-work}
\begin{figure*}[htb]
    \centering
    \includegraphics[width=0.9\textwidth]{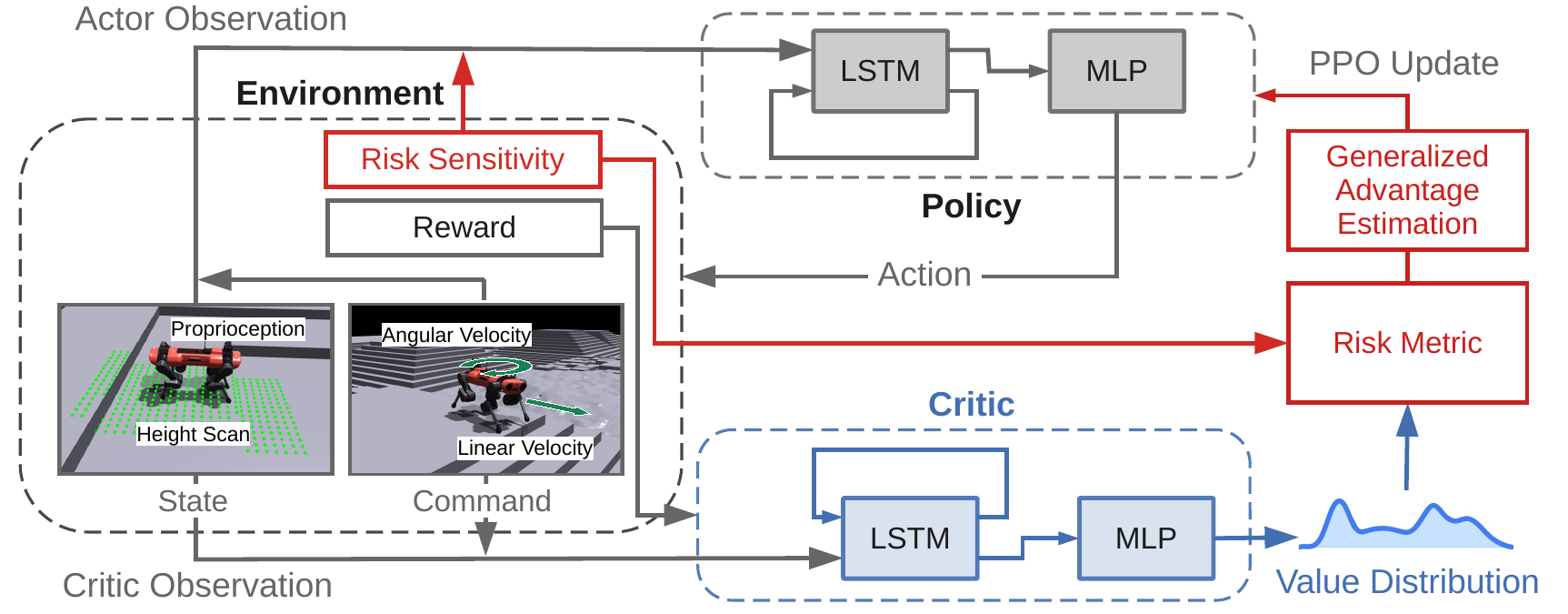}
    \caption{Architecture overview. The critic learns to predict a value distribution, used in combination with a risk metric to update the policy. The policy is conditioned on the risk parameter. The risk parameter is part of the command, set by the operator.}
    \label{fig:architecture}
    \vspace{-0.5cm}
\end{figure*}

\textbf{Legged Locomotion}\;\;
Legged locomotion is a thoroughly studied research area in robotics. Early successes have been achieved by model-based methods~\cite{anymal,mit-cheetah}. However, due to their limited ability to generalize to more complex and unpredictable environments, learned methods have recently gained interest. Several deep \gls{rl} algorithms have solved simulated legged locomotion tasks~\cite{trpo,ppo,ddpg,async-rl}. Beyond that, \gls{rl} policies learned in simulation have been successfully transferred to real-world scenarios, either through high simulation fidelity and data augmentation~\cite{learning-to-walk-in-minutes,legged-locomotion-egocentric-vision,learning-robust-perceptive-locomotion,legged-locomotion-dynamics-randomization,legged-locomotion-agile-sim-to-real,learning-quadrupedal-locomotion-over-challenging-terrain, hwangbo2019learning} or test-time adaptation~\cite{locomotion-keep-on-learning}.

\textbf{Distributional Reinforcement Learning (Dist. RL)}\;\;
\gls{drl} algorithms learn a value distribution, as defined by the recursive distributional Bellman Equation~\cite{c51}, which replaces the role of the value function in classical RL. Value-based \gls{drl} algorithms~\cite{c51,qrdqn,iqn,fqf} use this value distribution for value-iteration, commonly maximizing its expectation when comparing different actions. These methods differ primarily in how they represent the value distribution: as a categorical distribution \cite{c51}, by its cumulative distribution at fixed quantiles \cite{qrdqn}, or at arbitrary quantiles \cite{iqn,fqf}. Generally, algorithms with more accurate representations achieve better results. To extend \gls{drl} to continuous actions, the value distribution is used as the critic in an actor-critic architecture~\cite{d4pg,tqc,dsac,gmac,sdpg}. Several works explore \gls{drl} theoretically~\cite{drl-cramer,drl-regularization-perspective,drl-statistics-samples,drl-qtd}, use the additional information contained in the value distribution to improve exploration~\cite{drl-efficient-exploration}, or for risk sensitivity~\cite{iqn,dsac,drl-worst-cases-pg,drl-risk-sensitive-driving,drl-cvar-optimization,risk-sensitive-policy-dist-rl}. \gls{drl} has been used for algorithm discovery~\cite{drl-faster-matrix-multiplication} and as an AI racing agent in the Gran Turismo 7 game~\cite{gt-sophy}. Further, \gls{drl} control policies have been successfully employed in the real world: for robotic grasping in a lab environment~\cite{qt-opt}, robot soccer~\cite{agile-bipedal-soccer-skills}, and stratospheric balloon navigation~\cite{drl-balloon}.

\textbf{Safe Reinforcement Learning}\;\;
Safe \gls{rl} is a thoroughly studied research area~\cite{safe-rl-survey}. The work of \cite{drl-worst-cases-pg} is most closely related to ours, learning a risk-averse policy with online adaptation of risk-aversion. They model the value function as a Gaussian distribution and integrate it into an actor-critic architecture. When updating the policy, they maximize the Conditional Value at Risk (CVaR) metric and condition the policy on the confidence. \cite{automatic-risk-adaptation} extend this work to adjust risk sensitivity automatically. Further, \gls{dsac} is a state-of-the-art risk sensitive \gls{rl} algorithm that integrates a value distribution into the maximum entropy objective of \gls{sac}~\cite{sac}. Several works explore risk metrics in \gls{drl} with fixed risk parameters~\cite{iqn,drl-risk-sensitive-driving,drl-cvar-optimization,risk-sensitive-policy-dist-rl,qt-opt,urpi2021riskaverse}.
%
%
\section{Method}
\label{sec:method}
Our goal is to learn a locomotion policy \(\pi_\phi(a \vert s ; \beta) \) that maps the robot's state \(s\) to desired joint positions \(a\), is conditioned on a risk parameter \(\beta\), and has learned weights $\phi$. The state consists of proprioceptive measurements (base velocity, projected gravity, joint positions and velocities), the previous actions, samples from a local height map around the robot, and the input commands (desired linear and angular base velocities). The risk parameter is a scalar value and depends on the metric used during training. We use the robot dynamics model from \cite{learning-to-walk-in-minutes}, including learned actuator dynamics. An overview of the architecture can be found in Figure~\ref{fig:architecture}.

\textbf{Distributional Proximal Policy Optimization (DPPO)}\;\;
We base our \gls{rl} algorithm on the \gls{ppo} algorithm~\cite{ppo} as it shows superior performance for learning legged locomotion in our setup, compared to \gls{ddpg}~\cite{ddpg}, \gls{td3}~\cite{td3}, \gls{sac}~\cite{sac}, and \gls{dsac}~\cite{dsac} (Experiment A: Section~\ref{sec:experiment-algorithmic-performance}).

\textbf{Critic Representation}\;\;
As the distributional critic, we use the Quantile Regression DQN (QR-DQN)~\cite{qrdqn}. It parameterizes the value distribution as a uniform distribution supported on $\left\{ \theta_1(s), ..., \theta_N(s) \right\}$, which can be written as
\begin{equation}
\label{eq:quantile-distribution}
Z_\theta(s) = \frac{1}{N} \sum_{i=1}^N \delta_{\theta_i(s)},
\end{equation}
where $\delta_z$ is a Dirac at $z \in \mathbb{R}$. The support positions $\theta_i(s)$ are state-dependent and predicted by the critic's neural network.

\textbf{Critic Updates}\;\;
To compute reward-to-go value targets, we rely on the sample-replacement strategy $\text{SR}(\lambda)$~\cite{gmac}. This technique extends the temporal-difference $\text{TD}(\lambda)$ method~\cite{rl-intro} for computing multi-step value targets to \gls{drl}. As the name suggests, it computes value targets $\T Z_\theta(s)$ as a mixture of sample distributions. It starts with samples of the distribution at the final time-step $Z_\theta(s_T)$. Then, going backwards in time ($t = T-1, ..., 0$), the sample distribution is discounted, shifted by the obtained reward $r_t$, and a fraction $1 - \lambda$ of samples is replaced with samples of the distribution $Z_\theta(s_t)$. $1$-step and $N$-step target distributions are recovered by setting $\lambda = 0$ and $\lambda = 1$, respectively. Critic updates minimize the energy distance
\begin{equation}
\mathcal{L} = 2 \E_{i,j} \left[ \theta_i - \T\theta_j \right] - \E_{i,j} \left[ \T\theta_i - \T\theta_j \right] - \E_{i,j} \left[ \theta_i - \theta_j \right]
\end{equation}
between the $\text{SR}(\lambda)$ target distribution and samples from the predicted critic distribution $Z_\theta(s)$.
\begin{figure}[t]
    \centering
    \footnotesize
    \includegraphics[width=\linewidth]{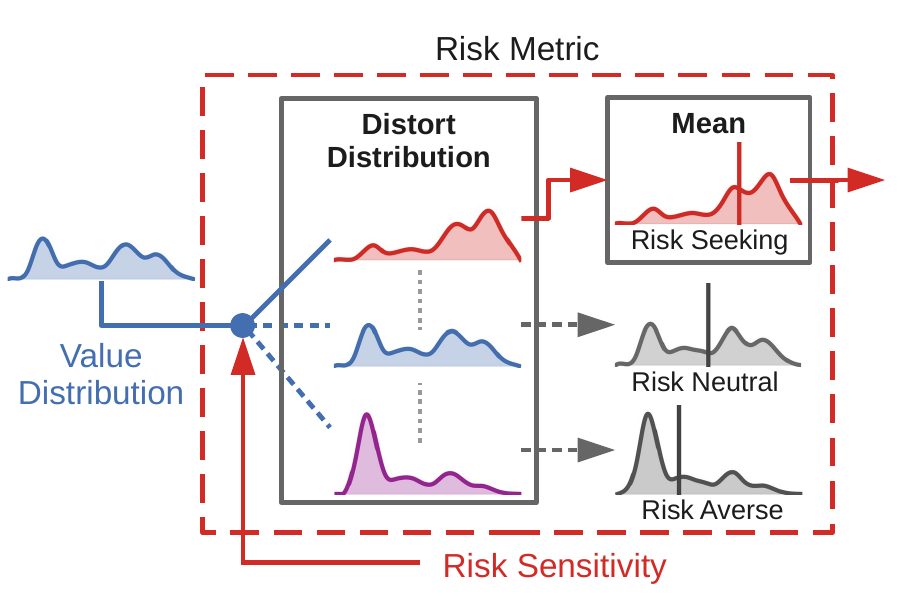}
    \caption{Application of a risk metric. The risk sensitivity selects how the value distribution is distorted. The mean value of the distorted distribution is provided for Generalized Advantage Estimation.}
    \label{fig:architecture-risk-metric}
    \vspace{-0.6cm}
\end{figure}

\textbf{Rewards}\;\;
We adapt the rewards from \cite{learning-to-walk-in-minutes}, which primarily reward command tracking and penalize colliding / expending energy. We introduce an  alive reward and modify the collision penalties to be impact force-dependent. Further, the joint motion and torque penalties are increased based on a curriculum.

\textbf{Risk Quantification}\;\;
To quantify the risk associated with different states, we apply a risk metric to the value distribution $Z_\theta(s)$, as depicted in Figure \ref{fig:architecture-risk-metric}. We thereby follow \cite{robot-risk-metrics} who argue for using distortion risk metrics, a class of metrics with properties sensible for robotics applications.
Specifically, we use the Wang~\cite{wang} metric, which encodes a subjective measure of risk. We further use Conditional Value at Risk (CVaR)~\cite{cvar}, which measures the expected worst-case return (beyond a confidence level).
For the distribution produced by the QR-DQN, we compute these metrics as the expectation under a distortion $g(\tau)$ of the distribution supports~\cite{dsac}. We denote the quantile fractions of the distribution as $(\tau_i)_{i=0,...,N}$ with $\tau_i = i/N$. The distorted expectation is computed as
\begin{equation}
V_\beta(s) = \int_0^1 g_\beta'(\tau) Z_\theta^\tau(s) d\tau = \sum_{k=1}^N (g_\beta({\tau}_k) - g_\beta(\tau_{k-1}))\theta_k(s),
\end{equation}
where we denote the $\tau$-quantile of the value distribution as $Z_\theta^\tau(s)$. We distort the quantile fractions for CVaR as
\begin{equation}
g_\beta^\text{CVaR}(\tau) = \min\left(\frac{\tau}{\beta}, 1.0\right),
\end{equation}
where $\beta$ is a scalar risk parameter. We choose $\beta = 1$ for risk-neutrality and $0 < \beta < 1$ for risk aversion. The CVaR metric, in effect, computes the expectation of the tail of the value distribution, where $\beta$ decides the cutoff. For the Wang metric, we compute the distorted quantile fractions as
\begin{equation}
g_\beta^\text{Wang}(\tau) = \Phi(\Phi^{-1}(\tau) + \beta),    
\end{equation}
where $\Phi$ is the standard normal distribution and $\beta$ is the scalar risk parameter. We choose $\beta = 0$ for risk-neutrality, $\beta > 0$ for risk aversion, and $\beta < 0$ for risk affinity.

For the CVaR metric, we train the policy on risk parameters in the range $\beta \in (0, 1]$ while for the Wang metric, we use $\beta \in [-1.5, 1.5]$. We find that the risk sensitive behavior generalizes well beyond the parameter bounds used during training. That is, using $\beta$ outside the training range results in stable but more risk-averse or affine behavior.

\textbf{Policy Updates}\;\;
We update the policy by maximizing the usual \gls{ppo} clip-objective
\begin{equation}
\label{eq:ppo-objective}
\begin{split}
& \mathcal{L} = \min\left( \frac{\pi_\phi(s \vert a; \beta)}{\pi_{\phi_\text{old}}(s \vert a; \beta)}A^{\pi_{\phi_\text{old}}}(s, a; \beta), g\left( \epsilon, A^{\pi_{\phi_\text{old}}}(s, a; \beta) \right) \right) \\
& \text{where} \;\;\; g(\epsilon, A) =
    \begin{cases}
      (1 + \epsilon)A, & \text{if $A \geq 0$};\\
      (1 - \epsilon)A, & \text{if $A < 0$}.
    \end{cases}
\end{split}
\end{equation}
When computing the advantages $A^{\pi}(s_t, a_t; \beta)$, we replace the value estimates with our risk sensitive value estimates. To estimate the advantage, we use a truncated version of \gls{gae}~\cite{gae} motivated by \cite{async-rl}. It approximates the advantage for each sample $(s_t, a_t, r_t, s_{t+1})$ by rolling out the policy $\pi(a \vert s ; \beta)$ for $T$ timesteps without looking beyond the $T$-th timestep. We compute truncated \gls{gae} estimates as
\begin{equation}
\label{eq:advantage-estimation}
\begin{split}
& A^\pi(s_t, a_t; \beta) = \sum_{l=0}^{T-t-1} (\lambda\gamma)^{l} \delta_{t+l}^r \\
& \text{where} \;\;\; \delta_t^r = r_t + \gamma V_\beta(s_{t+1}) - V_\beta(s_t)
\end{split}
\end{equation}
and $\lambda$ is the \gls{gae} bias / variance trade-off hyperparameter.


\section{Experiments}
\label{sec:experiments}

We evaluate our method through a series of experiments. In sections~\ref{sec:experiment-obstacle-course} to~\ref{sec:experiment-risk-sensitive-performance}, we conduct simulation experiments to study risk sensitivity and compare our method with other approaches. In Section~\ref{sec:hardware-results}, we conduct hardware experiments on the ANYmal robot~\cite{anymal}.


\begin{figure*}[htb]
    \centering
    \footnotesize
    \includegraphics[width=\linewidth]{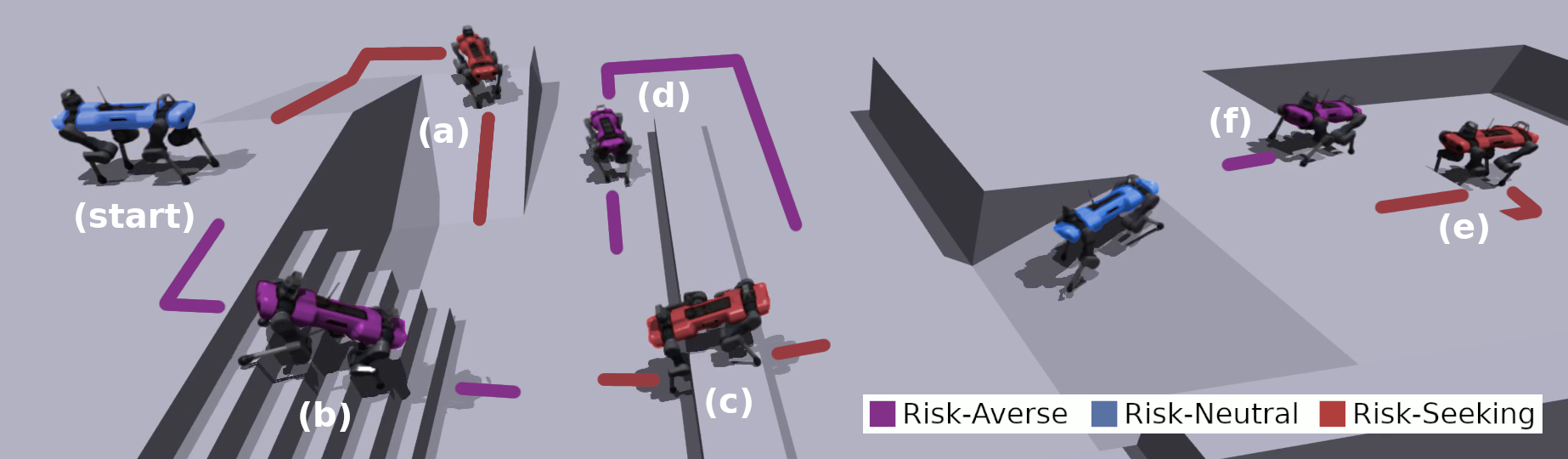}
    \caption{Robot operated remotely along an obstacle course. Depending on the chosen path, different risk sensitivities are preferable. For the incline along path~(a), a risk-seeking policy~(\redsquare{}) can be chosen for increased walking speed. Using a risk-averse policy~(\magentasquare{}) when descending the stairs along route~(b) ensures the robot's safety. A risk-averse policy~(\magentasquare{}) won't climb the dangerous obstacle~(c) and thus would have to walk around it, along route~(d). Meanwhile, setting the policy to risk-seeking~(\redsquare{}) allows the robot to surmount the obstacle~(c). To step down into the deep pit along route~(e), one must set the sensitivity to risk-seeking~(\redsquare{}). The risk-averse policy~(\magentasquare{}) will refuse to step into the pit~(f) as it may lead to a crash. Video: \href{https://youtu.be/GGFXpF4qeVY}{https://youtu.be/GGFXpF4qeVY}.}
    \label{fig:experiments-obstacle-course}
    \vspace{-0.5cm}
\end{figure*}

\textbf{Training Setup}\;\;
\label{sec:experiment-training-setup}
To train policies, we used the legged\_gym library~\cite{learning-to-walk-in-minutes}, based on the IsaacGym simulator. We used the same observation and action space, except for the additional risk sensitivity. The risk sensitivity is uniformly sampled with the command during training. The training environment comprises seven different terrains with ten difficulty levels per terrain. Robots spawn at the center of each terrain with a random offset.

\textbf{Simulation Experiment Setup}\;\;
The quantitative simulation experiments (Sections~\ref{sec:experiment-algorithmic-performance} and \ref{sec:experiment-risk-sensitive-performance}) were conducted in a deterministic environment to enable a fair comparison. In both experiments, we spawned 72 robots at the center of each environment tile with initial z-rotations uniformly distributed across $360^\circ$ and set the command to a forward velocity of $1\,\text{m/s}$.

\subsection{Simulation Experiment: Obstacle Course}
\label{sec:experiment-obstacle-course}

To demonstrate different risk sensitivities in a teleoperation scenario, we navigated a robot through the obstacle course depicted in Figure~\ref{fig:experiments-obstacle-course}. The shortest path from start to goal in this environment is to descend steep stairs (b) [$25$\unit{cm} steps], cross the barrier (c) [$55$\unit{cm} height], walk up the ramp, and descend into the pit (e) [$85$\unit{cm} depth]. To follow this route successfully, the operator could make use of the different risk sensitivities.

\textbf{Discussion}\;\;
Based on risk sensitivity, the robot exhibited different locomotion behaviors. When using a risk-seeking policy, the robot generally walked faster and attempted to overcome obstacles despite the risk of falling. When commanding the robot over the barrier along the path (c), the risk-seeking policy always attempted to follow the command. In most cases, it managed to surmount the barrier but, in a few cases, got stuck on the barrier or fell down the opposite side. When sending the risk-seeking policy along path (b), it sometimes managed to descend the stairs but fell in most cases. We obtained similar results when descending into the pit along the path (e), where the robot complied each time but often crashed. The risk-averse policy, on the other hand, demonstrated a slow but safe walking gate. It thus safely descended the stairs along path (b) on each attempt. However, the risk-averse policy refused to climb the barrier along the path (c) and descend into the pit (f).
This experiment showed that different risk sensitivities change the behavior of the robot. This allows either an operator or a navigation system to make use of the risk-sensitivity to increase safety during the mission. 

\subsection{Simulation Experiment: Algorithm Performance}
\label{sec:experiment-algorithmic-performance}

\begin{figure}[htb]
    \centering
    \footnotesize
    \includegraphics[width=\linewidth]{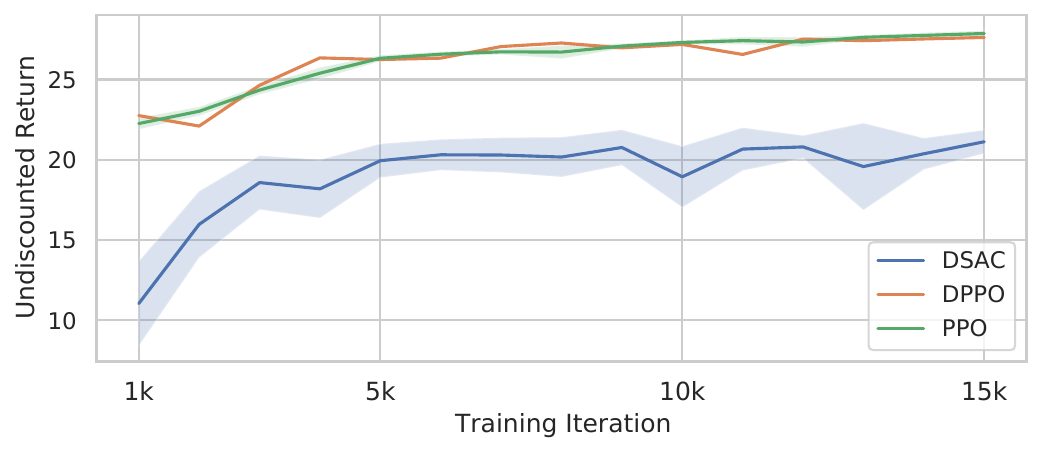}
    \caption{Average return in the evaluation environment. Shaded regions indicate $95\%$ confidence intervals across seeds and evaluation spawns. Hyperparameters for DPPO were not tuned.}
    \label{fig:experiments-algorithm-performance}
    \vspace{-0.2cm}
\end{figure}

We compare the attained return of our method with \gls{ppo}~\cite{ppo}, commonly used in state-of-the-art legged locomotion approaches~\cite{legged-locomotion-egocentric-vision,learning-robust-perceptive-locomotion}, and \gls{dsac}~\cite{dsac}, a recent distributional actor-critic algorithm that can learn risk sensitive policies. \gls{dsac} has previously been tested on legged locomotion~\cite{automatic-risk-adaptation} and, with some adaptations, been employed in complex control scenarios~\cite{gt-sophy}. For each method, we trained $10$ policies. In this experiment, our method was trained with ``risk-neutral'' expectation instead of a risk metric. Due to computation constraints, we employed \glspl{mlp} for both actor and critic across all three methods. Figure~\ref{fig:experiments-algorithm-performance} shows the learning curve for each method.

\textbf{Discussion}\;\;
We found that both DPPO (return of $27.62$ after $20$k~iterations) and \gls{ppo} ($27.86$) significantly outperformed \gls{dsac} ($21.1$). Qualitatively, \gls{dsac}, which also uses a Dist. RL formulation, failed to learn a proper locomotion policy. Despite reusing the hyperparameters from \cite{learning-to-walk-in-minutes}, extensively tuned for \gls{ppo}, our method performs on-par with \gls{ppo}. We thus establish our method as competitive in terms of attained return. We highlight that the core contribution of our work is not outperforming \gls{ppo} in terms of attained return, but to allow for risk sensitive behavior.

\subsection{Simulation Experiment: Risk Sensitive Performance}
\label{sec:experiment-risk-sensitive-performance}
\begin{figure*}[htb]
    \centering
    \footnotesize
    \begin{subfigure}{0.33\linewidth}
        \includegraphics[width=\linewidth]{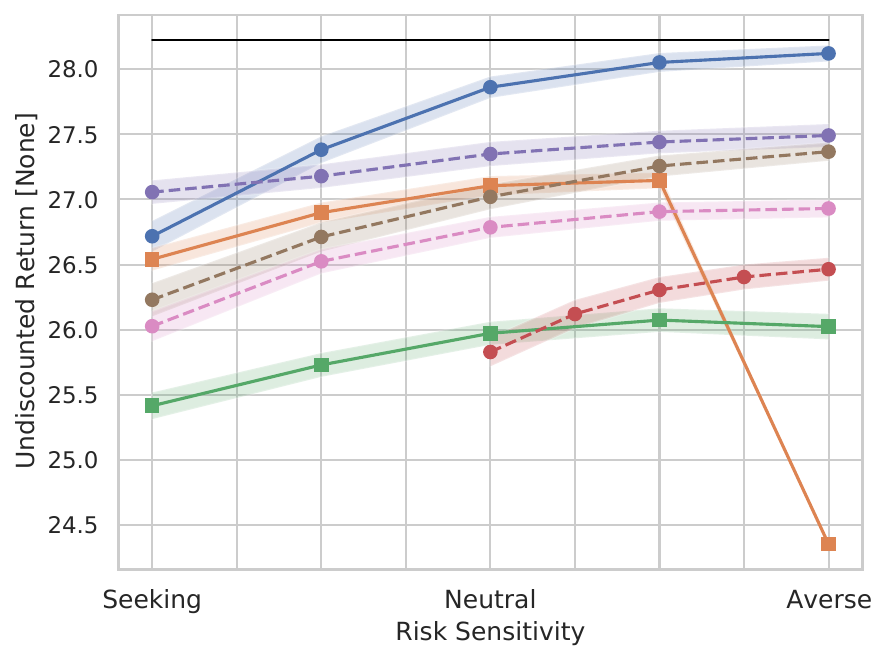}
    \end{subfigure}
    \begin{subfigure}{0.33\linewidth}
        \includegraphics[width=\linewidth]{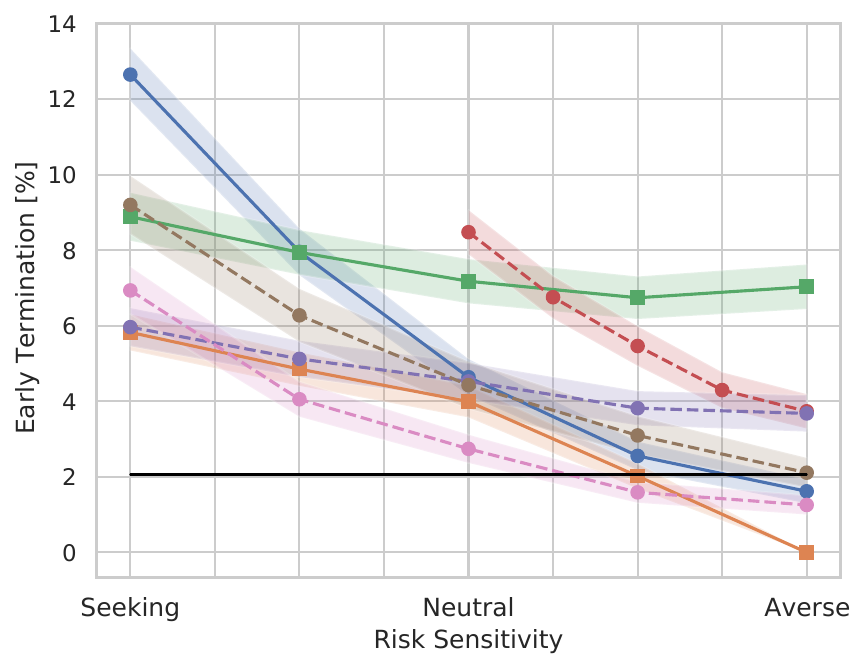}
    \end{subfigure}
    \begin{subfigure}{0.33\linewidth}
        \includegraphics[width=\linewidth]{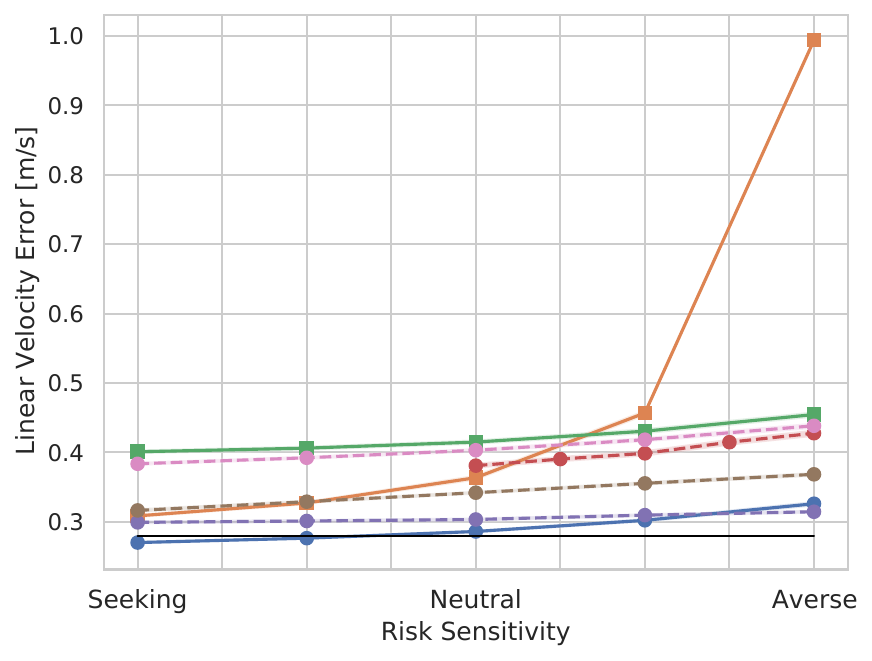}
    \end{subfigure}
    \begin{subfigure}{\linewidth}
        \includegraphics[width=\linewidth]{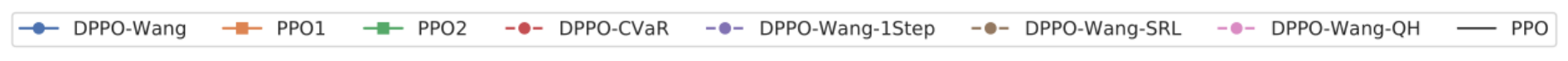}
    \end{subfigure}
    \caption{Average undiscounted return, fraction of episodes in which the robot terminates before episode end, and $L2$ distance between commanded and actual linear velocity. PPO baseline is included for reference and does not change with risk sensitivity. Shaded regions show $95\%$ confidence intervals across evaluation spawns. Only a single seed was trained for each method due to computation constraints. Methods are listed in table~\ref{tbl:ablation-study}.}
    \label{fig:experiments-risk-sensitive-performance}
    \vspace{-0.5cm}
\end{figure*}
\begin{table}[htb]
  \centering
  \scriptsize
  \begin{tabularx}{\linewidth}{lll}
    \toprule
    \textbf{Name} & \textbf{Loss Function} & \textbf{Value Target} \\
    \midrule
    DPPO-Wang & Energy & $N$-step distribution [$\text{SR}(\lambda)$, $\lambda = 1$] \\
    \addlinespace
    DPPO-Wang-QH & Quantile Huber & $N$-step distribution\\
    \addlinespace
    DPPO-Wang-SRL & Energy & $\text{SR}(\lambda)$, $\lambda = 0.95$ \\
    \addlinespace
    DPPO-Wang-1Step & Energy & 1-step distribution [$\text{SR}(\lambda)$, $\lambda = 0$] \\
    \bottomrule
  \end{tabularx}
  \caption{Overview of ablated components on DPPO-Wang.}
  \label{tbl:ablation-study}
  \vspace{-0.3cm}
\end{table}

We further compare the attained return between different ablations of our method (listed in Table~\ref{tbl:ablation-study}) as well as non-distributional reward-shaping baselines, introduced below.
Return is computed using the original reward formulation. We show the undiscounted return in Figure~\ref{fig:experiments-risk-sensitive-performance}. We also include the fraction of early terminations and the linear velocity tracking error. These two metrics underline our qualitative assessment on changing behavior with risk sensitivity from Section~\ref{sec:experiment-obstacle-course} with quantitative results. 

\textbf{PPO Baseline}\;\;
To contrast our distributional approach, we implement \gls{ppo} baselines that modulate ``risk''-preference through the reward formulation. In these baselines, risk-preference is introduced before accounting for environment dynamics, setting them apart from our \gls{drl} approach.
Our first approach, which we call PPO1, adapts individual terms of the reward formulation. Specifically, we scale the linear velocity tracking reward by $\beta \in [0, 2]$. This scaling aims at trading-off command tracking and other rewards, most notably the alive reward. The policy should thus learn, depending on the risk parameter, to mediate differently between command compliance and keeping the robot safe.

However, reward tuning is a complex process. It is often unclear how a reward term affects the learned policy. Further, individual terms cannot be tuned in isolation as they influence one another: e.g. high torque penalties may prevent the algorithm from discovering the linear velocity tracking reward. The approach of the PPO1 baseline further increases this design complexity (especially when scaling multiple reward terms). Thus, in a second approach we call PPO2, we rescale the full reward formulation in a principled manner. We treat individual reward terms as Dirac impulses defining a probability distribution (mirroring Equation~\ref{eq:quantile-distribution}) and compute the distorted mean by applying the Wang metric. As risk parameter, we choose $\beta \in [-0.25, 0.25]$ as we are unable to learn a policy for a larger range.

The policy $\pi(a \vert s; \beta)$ learned by either baseline is conditioned on the scalar parameter $\beta$, emulating the risk parameter from our method. In both methods, the critic predicts the expected return used for \gls{gae} computation as in standard \gls{ppo}. Aside from the critic, we use the same training setup, rewards, and hyperparameters as for DPPO. We expect the PPO1 baseline to learn risk-averse behavior for $\beta \in [0, 1)$ and risk-seeking behavior for $\beta \in (1, 2]$. It reverts to the initial, ``risk-neutral'', reward formulation for $\beta = 1$. We expect the PPO2 baseline to exhibit behavior appropriate for the Wang metric (see Section~\ref{sec:method}).

\textbf{Discussion}\;\;
For a policy to qualify as risk sensitive, we expect the Early Terminations and the Linear Velocity Error to change appropriately across different risk settings. All methods showed such risk sensitive behavior that matches our intuition: risk-seeking policies had improved command tracking but also more premature terminations; risk-averse policies sacrificed command tracking to keep the robot safe.

The return generally increased with risk-averseness. This trend could be due to high-reward actions which, despite often failing, would still be pursued by a risk-seeking policy. In our environment, however, such risk-seeking behavior was not rewarded.
Our method significantly outperformed the reward-shaping baselines regarding return. Especially the PPO2 baseline had severely lower returns. These mirrored its high early termination rate and linear velocity tracking error.
For the PPO1 baseline, we experienced outlier behavior for the most risk-averse setting: since the velocity tracking reward was set to zero during training, the robot learned to stop walking. The undesired behavior of PPO1 and low return of PPO2 demonstrate the difficulties when incorporating risk sensitivity through the standard \gls{rl} framework.  

Comparing our method to its ablations, none attained similar return. DPPO-CVaR most significantly underperformed in our experiments. This underperformance may be due to the CVaR metric disregarding an entire portion of the value distribution, appearing to result in less stable training. Thus, our design decisions are experimentally justified.


\subsection{Hardware Experiment}
\label{sec:hardware-results}

\begin{figure*}[htb]
    \centering
    \includegraphics[width=1.0\linewidth]{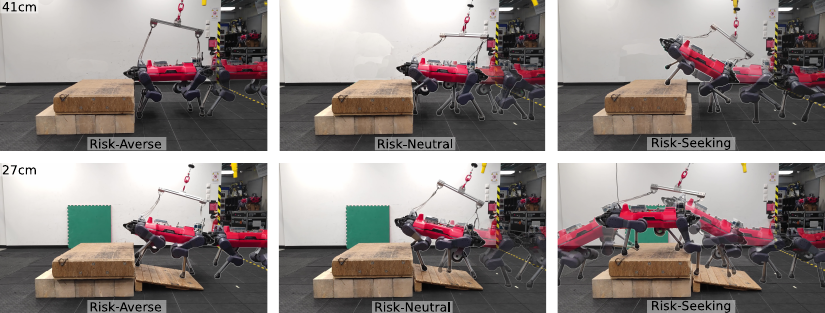}
    \caption{ANYmal showed different step-up behaviors depending on risk sensitivity and step height. Video: \href{https://youtu.be/GGFXpF4qeVY}{https://youtu.be/GGFXpF4qeVY}.}
    \label{fig:experiments-hardware}
    \vspace{-0.5cm}
\end{figure*}

Our learned policy transfers to the real world using appropriate domain randomization~\cite{learning-to-walk-in-minutes}. We conducted experiments on the ANYmal robot~\cite{anymal} with onboard perception~\cite{mikielevation2022} to replicate the risk sensitive behavior observed in simulation.
We commanded the robot to walk up steps of height $27$\unit{cm} and $41$\unit{cm} with a speed of $1.2\,\text{m/s}$ and counted \textit{refusals}, \textit{failures}, and \textit{successes} to surmount the obstacle. A high step has some risk of the robot falling and, thus, a risk-averse policy might avoid it while a risk-seeking policy should neglect the risk.
The experiments are depicted in Figure \ref{fig:experiments-hardware}. We collected a total of $45$ tries, $8$ for each policy on the $41$\unit{cm} step height and $7$ on the $27$\unit{cm} step height. The results are listed in Table \ref{tbl:experiments-hardware}.

\begin{table}[htb]
  \scriptsize
  \centering
  \begin{tabularx}{\linewidth}{l|YYY|YYY}
    \toprule
    & \multicolumn{3}{c|}{$27 cm$ Step} & \multicolumn{3}{c}{$41 cm$ Step}\\
    \midrule
    \textbf{Risk Sensitivity} & \textbf{Refusal} & \textbf{Failure} & \textbf{Succ.} & \textbf{Refusal} & \textbf{Failure} & \textbf{Succ.}\\
    \midrule
    Averse ($\beta = -1.5$) & $25\%$ & $50\%$ & $25\%$ & $85.7\%$ & $14.3\%$ & $0\%$  \\
    \addlinespace
    Neutral ($\beta = 0.0$) & $12.5\%$ & $75\%$ & $12.5\%$ & $0\%$ & $100\%$ & $0\%$ \\
    \addlinespace
    Seeking ($\beta = +1.5$) & $0\%$ & $37.5\%$ & $62.5\%$ & $0\%$ & $100\%$ & $0\%$ \\
    \bottomrule
  \end{tabularx}
  \caption{Hardware experiment results. The robot received a $1.2\,\text{m/s}$ command to walk up the step. We counted the number of \textit{refusals}, \textit{failures} (active attempts to surmount the obstacle without ultimate success), and \textit{successes}. The given percentages indicate how often each behavior was observed during the experiments.}
  \label{tbl:experiments-hardware}
  \vspace{-0.5cm}
\end{table}

\textbf{Discussion}\;\;
The robot's actions aligned well with the risk sensitivity of its assigned policies. When commanded to be risk-averse, the robot hesitated before climbing the step while it tried to follow with a risk-seeking command. On the higher step height ($41$\unit{cm}), the risk-averse policy mostly refused to walk the step. We often observed the policy standing before the obstacle, unresponsive to its forward command. The risk-seeking and neutral policies, while failing to surmount the obstacle as well, exhibited more aggressive behavior. The difference between the risk-neutral and risk-seeking policies became emphasized for the smaller step size ($27$\unit{cm}). In this experiment, the risk-seeking policy complied every time, often surmounting the obstacle. The risk-neutral policy mostly attempted to overcome the obstacle but failed. A slight inconsistency arose between the risk-neutral and risk-averse policies, where the latter succeeded in walking the step more often. Overall, the risk-neutral policy attempted to walk the step more often than the risk-averse policy. These observations mirror our findings in simulation and confirm that risk sensitivity successfully transfers to hardware.
%
%
%



\section{Conclusion}
\label{sec:conclusion}

Our work presents a Dist. RL approach for learning risk sensitive locomotion policies. These policies allow the robot to adapt its behavior to environment risks. A risk preference is encoded through a single parameter that can be modulated during deployment. To incorporate risk sensitivity into the \gls{rl} formulation, we introduce risk sensitive advantage estimates.
The emergent risk sensitive behavior, demonstrated in simulation and on hardware, opens up a new application domain for \gls{drl} in real-world applications. Our work directly benefits teleoperation and navigation, given that a user / planner can now change the risk affinity of the robot during deployment, increasing safety while retaining full locomotion capability.
Further research is needed to establish procedures on evaluating risk sensitivity and study risk sensitive \gls{drl} from a theoretical standpoint: how the choice of risk metric impacts the resulting policy and how well the estimated value distribution matches the true return distribution are some suggestions. Additionally, integrating DPPO into a navigation system that utilizes risk sensitivity is promising for future work.

\bibliographystyle{IEEEtran}
\balance
\bibliography{IEEEabrv,references}

\vfill
\end{document}

%% file: preamble.tex
\usepackage{times}
\usepackage{amsmath,amsfonts}
\usepackage{amssymb}
\usepackage{algorithmic}
\usepackage{algorithm}
\usepackage{array}
\usepackage{booktabs} 
\usepackage{textcomp}
\usepackage{stfloats}
\usepackage{url}
\usepackage{verbatim}
\usepackage{graphicx}
\usepackage{todonotes}
\usepackage{tabularx}
\usepackage{mathtools}
\usepackage{multicol}
\usepackage[bookmarks=true]{hyperref}
\usepackage{annotate-equations} 
\usepackage{balance}
\usepackage{lipsum} 
\usepackage{siunitx}

\usepackage[abbreviations]{glossaries-extra}
\hypersetup{hidelinks}

\usepackage{subcaption}
\usepackage{etoolbox} 

\usepackage{siunitx} 
\def\eqref#1{Eq.~(\ref{#1})}

\glssetcategoryattribute{acronym}{indexonlyfirst}{true}

\definecolor{PaperOrange}{RGB}{251, 151, 39}
\definecolor{PaperMagenta}{RGB}{150, 36, 145}
\definecolor{PaperBlue}{RGB}{67, 110, 176}
\definecolor{PaperCyan}{RGB}{66, 173, 187}
\definecolor{PaperRed}{RGB}{210, 43,38}

\newcommand{\redsquare}{{\textcolor{PaperRed}{$\blacksquare$}}}

\newcommand{\magentasquare}{{\textcolor{PaperMagenta}{$\blacksquare$}}}